\pdfoutput=1
\documentclass[11pt]{article}

\usepackage{EMNLP2022}

\usepackage{times}
\usepackage{latexsym}

\usepackage[T1]{fontenc}

\usepackage[utf8]{inputenc}

\usepackage{microtype}

\usepackage{inconsolata}

\usepackage{graphicx}

\usepackage{tabularx}
\usepackage{multirow}
\usepackage{booktabs}
\usepackage{diagbox}
\usepackage{csquotes}

\usepackage{amsmath}

\usepackage{amssymb}

\usepackage{balance}

\title{Do Vision-and-Language Transformers Learn Grounded Predicate-Noun Dependencies?}

\author{
    Mitja Nikolaus\textsuperscript{1,2} \\ 
    \texttt{mitja.nikolaus@univ-amu.fr} \\\AND
   Emmanuelle Salin\textsuperscript{1} \and Stephane Ayache\textsuperscript{1} \and Abdellah Fourtassi\textsuperscript{1} \and Benoit Favre\textsuperscript{1} \AND
\normalfont \textsuperscript{1}Aix Marseille Univ, Université de Toulon, CNRS, LIS, Marseille, France \\
  \textsuperscript{2}Aix-Marseille Univ, CNRS, LPL, Aix-en-Provence, France \\
  }

\begin{document}
\maketitle
\begin{abstract}
Recent advances in vision-and-language modeling have seen the development of Transformer architectures that achieve remarkable performance on multimodal reasoning tasks.
Yet, the exact capabilities of these black-box models are still poorly understood. While much of previous work has focused on studying their ability to learn meaning at the word-level, their ability to track syntactic dependencies between words has received less attention.\\
We take a first step in closing this gap by creating a new multimodal task targeted at evaluating understanding of predicate-noun dependencies in a controlled setup.
We evaluate a range of state-of-the-art models and find that their performance on the task varies considerably, with some models performing relatively well and others at chance level. In an effort to explain this variability, our analyses indicate that the quality (and not only sheer quantity) of pretraining data is essential. Additionally, the best performing models leverage fine-grained multimodal pretraining objectives in addition to the standard image-text matching objectives.\\
This study highlights that targeted and controlled evaluations are a crucial step for a precise and rigorous test of the multimodal knowledge of vision-and-language models.
\end{abstract}

\section{Introduction}

\begin{figure}[t]
    \centering
    \includegraphics[width=.49\textwidth]{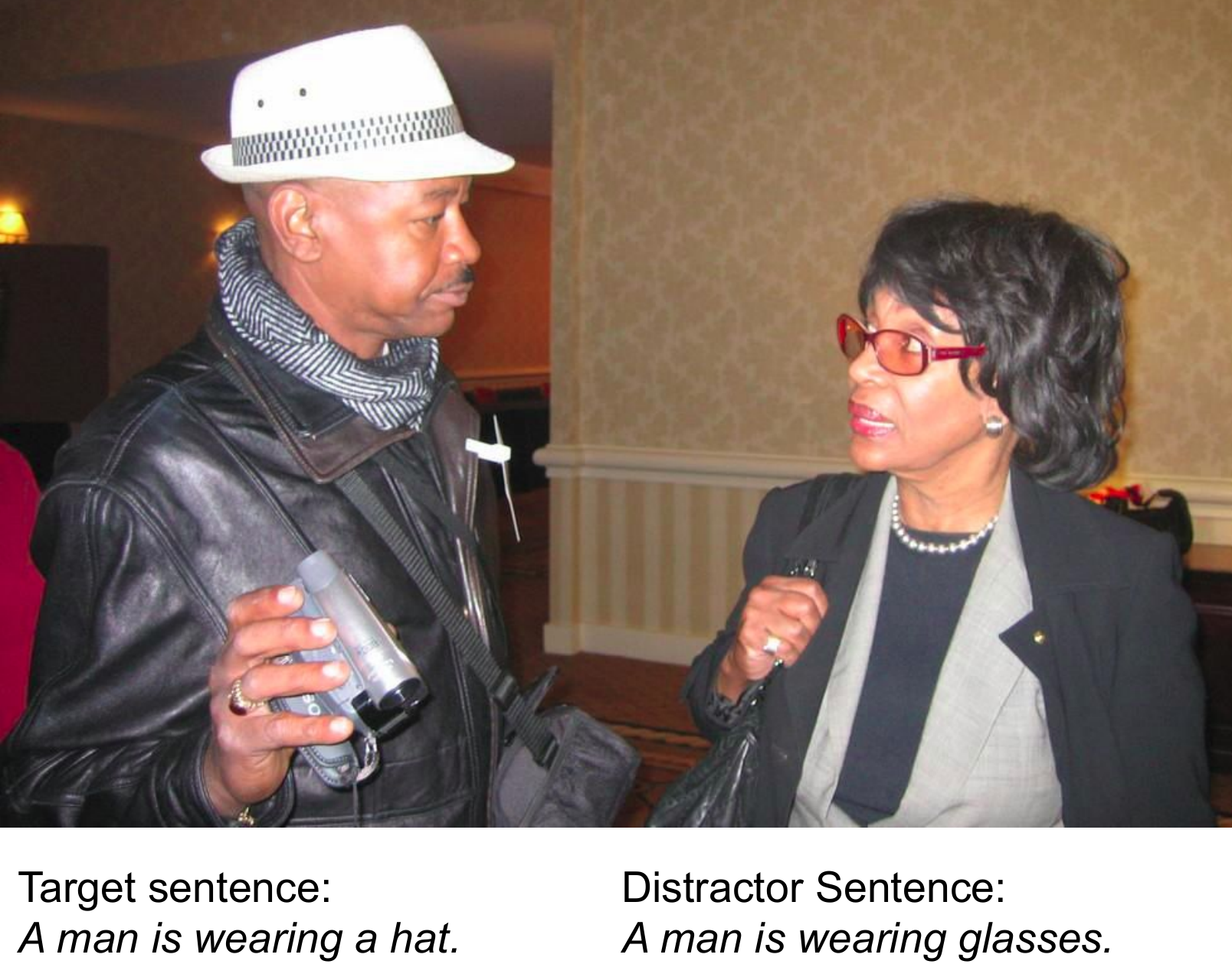}
    \caption{We evaluate V\&L models on their ability to track predicate-noun dependencies that require a joint understanding of the linguistic and visual modalities. The task is to find the correct sentence (choosing between the target and distractor) that corresponds to the scene in the image. In this example, the models should connect the predicate ``is wearing a hat'' to ``man''. A model that does not track dependencies would judge the distractor sentence ``A man is wearing glasses'' as equally likely, as there is a man is the image, as well as a person that is wearing glasses.}
    \label{fig:task}
\end{figure}

Vision-and-language (V\&L) models have recently shown substantial improvement on a range of multimodal reasoning tasks. Taking inspiration from successes in text-only Natural Language Processing \cite{devlin_bert_2019,brown_language_2020}, state-of-the-art V\&L models are usually composed of a Transformer-based architecture pre-trained in a self-supervised manner on large-scale data, and then fine-tuned on downstream tasks.

%This approach can be used to automatically describe images \cite{bernardi_automatic_2016}, answer questions regarding the image content \cite{antol_vqa_2015,goyal_making_2017,hudson_gqa_2019}, and reason about image content using natural language more generally \cite{zellers_recognition_2019,suhr_corpus_2019}.
%link referring expressions to image content \cite{kazemzadeh_referitgame_2014,mao_generation_2016},

While these models show remarkable performance on a range of tasks, more controlled and systematic analyses are necessary in order to obtain a better understanding of their exact multimodal knowledge. 

A range of studies has investigated their ability to map words to their visual referents for nouns \cite{kazemzadeh_referitgame_2014,mao_generation_2016,shekhar_foil_2017} and verbs \cite{ronchi_describing_2015,yatskar_situation_2016,pratt_grounded_2020,hendricks_probing_2021}, but there are only a few studies on whether recent V\&L models can capture multimodal syntactic dependencies between words and concepts.
%\footnote{However see \citet{akula_words_2020,thrush_winoground_2022} who created challenge datasets that test a model's ability to track word order, a simplified version of sentence-level dependencies.}

In this paper, we explore how well V\&L models learn predicate-noun dependencies across modalities (see example in Figure \ref{fig:task}).
To this end, we create an evaluation set that contains carefully selected images and pairs of sentences with minimal differences.
Given an image and two predicate-noun sentences, the models need to find the correct sentence corresponding to the image. Crucially, they can only succeed by taking into account the dependencies between the visual concepts in the image corresponding to the noun and predicate in the sentence.

As it has been shown that visual reasoning performance in several tasks can be spuriously augmented by capitalizing on textual biases in the training data \cite{goyal_making_2017,agrawal_dont_2018,hendricks_women_2018,cao_behind_2020}, we counter-balance our evaluation dataset in a way that controls for such exploitation of linguistic biases.
%Indeed, a controlled setup as presented in \citet{thrush_winoground_2022} is necessary to limit the impact of text bias and correctly evaluate multimodal ability.

%it is only possible to correctly match a sentence and an image by taking into account the dependency between the visual concepts in the image (corresponding to the noun) and the predicate in the sentence 

We evaluate pre-trained state-of-the-art V\&L models in a zero-shot setting and find that the ability to track predicate-noun dependencies varies considerably from model to model. Of all models tested, UNITER \cite{chen_uniter_2019} and LXMERT \cite{tan_lxmert_2019} show the highest scores, but their performance is still far from optimal. Other models such as ViLBERT \cite{lu_vilbert_2019} and CLIP \cite{radford_learning_2021} perform at chance level. We discuss how differences in the models  could explain their performance variability, highlighting the role of pretraining data quality and fine-grained multimodal pretraining objectives.

%Using a control task, we additionally estimate how much the performance of a modal in the original task is affected by its ability to detect the relevant nouns and predicates in an image (in isolation).
%disentangle the ability of a model to track predicate-noun dependencies with 

%Our results suggest that models trained on higher amounts of pretraining data do not necessarily perform better on the task, and that it is rather the quality than the quantity of pretraining data that matters. Further, the use of additional multimodal pretraining objectives other than image-text matching helps to learn more fine-grained multimodal dependencies.
%Models that leverage more recent image encoders do not perform better on the task, however they get better results on the control task. More targeted experiments are necessary to obtain more conclusive results regarding the role of image features.

Code to reproduce the analyses and run the evaluation on new models is publicly available at \url{https://github.com/mitjanikolaus/multimodal-predicate-noun-dependencies}.

\section{Related Work}

\paragraph{Targeted evaluation of V\&L models}

Recently, a growing number of tasks have been created for targeted evaluation of V\&L models' abilities to perform various multimodal reasoning.

%MS-COCO actions \cite{ronchi_describing_2015} specifically evaluates the understanding of actions, as well as involved subjects and objects. The task of situation recognition consists of producing structured summaries of images including the primary activity, as well as the semantic roles of involved the entities \cite{yatskar_situation_2016,pratt_grounded_2020}.
%visual semantic role labeling \cite{gupta}

%, by generating difficult challenge datasets. %using wrong captions with a strong similarity to the image, and more complex images to better study compositionality.
\citet{shekhar_foil_2017} create sets of distractor captions to analyze whether V\&L models are sensitive to single word replacements (with a focus on nouns).
Similar targeted evaluation datasets have also been proposed for referring expressions \cite{chen_cops-ref_2020}, image-sentence matching \cite{hu_evaluating_2019}, and Visual Question Answering \cite[VQA;][]{bogin_covr_2021}, with a focus on compositional reasoning.

Tasks such as visual semantic role labeling or situation recognition, typically involve classifying the primary activity depicted in an image, as well as the semantic roles of involved entities \cite{ronchi_describing_2015,lu_visual_2016,chao_hico_2015,gupta_visual_2015,yatskar_situation_2016,pratt_grounded_2020}.
While these studies demonstrate that V\&L models can learn semantic roles to some degree in a supervised learning setup, such tasks do not allow for a controlled evaluation of models in a zero-shot setting.

%\citet{chen_cops-ref_2020} proposes referring expression tasks that involve visually similar distractors, that more precisely evaluate compositional understanding. 
%\citet{bogin_covr_2021} tests for visually grounded compositional generalization, using a visual question answering dataset that requires complex reasoning over multiple images.
%In \citet{hu_evaluating_2019}, models are queried to select the image matching a sentence, given one of two semantically highly similar images.

In \citet{hendricks_probing_2021}, the authors evaluate state-of-the-art V\&L models in a controlled zero-shot setup and find that they still have more trouble understanding verbs compared to subjects or objects. They also observe that models trained on larger datasets with less descriptive captions perform worse than models trained on smaller, manually-annotated datasets.

Several works have also tried to shed more light on the precise multimodal semantic capabilities of V\&L models using probing techniques.
\citet{salin_are_2022} show that although state-of-the-art V\&L models can grasp some multimodal concepts such as color, they still do not fully understand more difficult concepts such as object size and position in the image. 
\citet{parcalabescu_seeing_2021} use probing to demonstrate that such models still lack the capability to correctly count entities in an image.

%\cite{dahlgren_lindstrom_probing_2020}: probing for counting, objects

\paragraph{Evaluation of grounded syntax}

\citet{akula_words_2020} tests for sensitivity to word order in referring expressions.
Similarly, \citet{thrush_winoground_2022} studies the ability of V\&L models to take word order into account by designing adversarial examples that require differentiating between similar image and text pairs, while the text pairs only differ in their word order. Their results suggest that state-of-the art models still lack precise compositional reasoning abilities.

\citet{li_what_2020} studies so-called \textit{syntactic grounding} of VisualBERT. They show that certain attention heads of the transformer architecture attend to entities that are connected via syntactic dependency relationships. 
However, such probing experiments do not necessarily indicate to what degree a model is actually \textit{using} the encoded information when making predictions.

In our work, we test a range of state-of-the-art models specifically on their ability to track predicate-noun dependencies. Crucially, we test the models in a much more controlled setting compared to previous work: Our setup involves visual distractors as well as control task, disentangling the challenge of understanding syntactic dependencies from more simple object and predicate recognition. Additionally, we strictly control for any possible linguistic bias by counter-balancing all evaluation examples.
%To achieve a high performance, a model needs to capture the dependencies between nouns and predicates in the image and correctly map it to the corresponding sentence.

\section{Methods}

\subsection{Evaluation Dataset}

%\paragraph{Evaluating predicate-noun dependencies}
We construct an evaluation dataset that is suited for evaluating the sensitivity to visually grounded predicate-noun dependencies in a zero-shot setup.

\begin{figure}[ht]
    \centering
    \includegraphics[width=.5\textwidth]{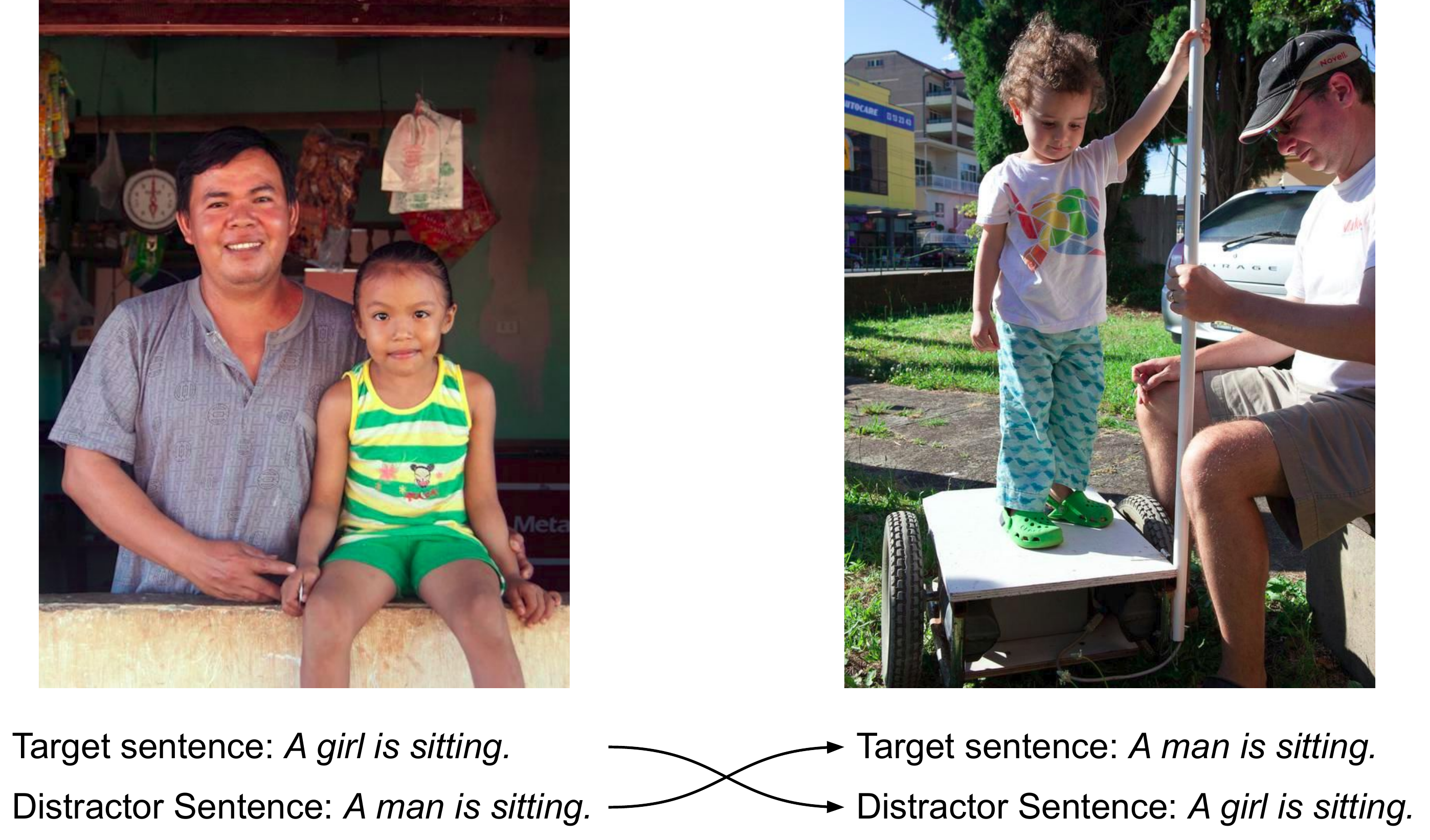}
    \caption{Counter-balanced evaluation: Each triplet has a corresponding counter-example, where target and distractor sentence are flipped.}
    \label{fig:counter_balancing}
\end{figure}

The data consists of pairs of triplets, and each triplet consists of an Image $I$, a target sentence $S_1$, and a distractor sentence $S_2$. Target and distractor sentences are minimal pairs, i.e. one sentence differs from the other only with regard to either the noun (e.g., ``A girl is sitting.'' vs. ``A man is sitting.'', Figure \ref{fig:counter_balancing}) or the predicate (e.g., ``A man is wearing a hat.'' vs. ``A man is wearing glasses.'', Figure \ref{fig:task}).

Crucially, the images always contain visual distractors, meaning that both the noun and the predicate of the distractor sentence are present in the image, but they do not have a noun-predicate relationship (e.g., for the distractor sentence ``A man is wearing glasses'', there is a man in the image, who is not wearing glasses, and a person wearing glasses, who is not a man).
Thus, it is necessary to take into account the \textit{dependency} between noun and predicate to distinguish between the target and distractor sentence (Figure~\ref{fig:task}).

\paragraph{Controlling for linguistic biases}
V\&L models have shown to rely sometimes on textual bias instead of using visual information \cite{goyal_making_2017,agrawal_dont_2018,hendricks_women_2018,cao_behind_2020}. 
For example, if a training dataset contains more often the phrase ``a girl is sitting'' than ``a man is sitting'', a model might prefer the caption ``a girl is sitting'' during evaluation only based on linguistic co-occurrence heuristics, irrespective of the visual content.
In our evaluation dataset, we control for potential linguistic biases in the training datasets by pairing every triplet with a corresponding counter-balanced example where target and distractor sentence are flipped. More specifically, for every triplet $(I_1, S_1, S_2)$, there exists a corresponding triplet $(I_2, S_2, S_1)$, as depicted in Figure \ref{fig:counter_balancing}. In that way, a model that does not take into account the visual modality cannot succeed in the task \cite[see also][]{nikolaus_evaluating_2021}.

\paragraph{Automatic pre-filtering}
Our evaluation dataset is based on Open Images \cite{kuznetsova_open_2020}. 
We pre-filter the images based on existing human-annotated object and relationship labels and bounding boxes. The objects refer to persons, animals, as well as inanimate objects. The relationships can either describe an action that an object is engaged in (e.g., \textsc{woman} - \textsc{sit}), or an action linking two objects (e.g., \textsc{man} - \textsc{wear} - \textsc{glasses}). All nouns in the selected relationships for our dataset refer to persons, due to lack of sufficient annotations for other kinds of agents.

We look for images that contain a target object-relationship pair as well as a distractor object-relationship pair for which either the target and distractor object are the same, but the relationships differ, or vice versa (as in the example in Figure \ref{fig:task}).
Additional details on the pre-filtering can be found in Appendix \ref{sec:app:pre_filtering}.

\paragraph{Manual selection}
We manually select suitable images after the automated pre-filtering, in order to ensure high quality of each example and in particular to verify that the distractor sentences are indeed incorrect given the images. This step is crucial, because many of the annotations in Open Images are incomplete, and an image may contain, for example, a woman that is sitting but not annotated as such (in this case, we disregard the image for our evaluation set). 
%Thus, in order to obtain clean evaluation examples, manual verification and selection is necessary.

We select pairs of examples and counter-examples and ensure that there are no duplicate images within the set of images for each object-relationship pair.

\paragraph{Sentence generation}
We generate target and distractor sentences based on the verified object and relationship annotations from Open Images.

We construct English sentences using a template-based approach. Given an object and a relationship, we add the indefinite article (a/an) in front of each noun and use all verbs in present progressive tense as this is most frequent in image-text datasets.\footnote{In cases where multiple connecting predicates between a verb and a noun are plausible (e.g. ``a man wearing glasses'' vs. ``a man with glasses''), we choose the construction that occurs most frequently in the Conceptual Captions training data \cite{sharma_conceptual_2018}. This dataset is most commonly used for training V\&L transformers.}
For example, from \textsc{woman} - \textsc{is} - \textsc{sit} we generate ``a woman is sitting.''; and from \textsc{man} - \textsc{hold} - \textsc{camera} ``a man is holding a camera.''.

This template-based approach is necessary for our controlled evaluation. As the choice of the exact template for the construction of the sentences may influence the results\footnote{For example, \citet{ravichander_systematicity_2020} found that results of some probing experiments can vary substantially with slight changes in wording.}, we evaluate the models, additionally, using a slightly different template, and we show that the overall result patterns remain largely similar (see Appendix \ref{sec:app:linguistic_robustness}).

\paragraph{Final evaluation set}

The final evaluation set contains 2584 triplets. For 1486 of these triplets, the distractor sentence contains an incorrect predicate and for the other 1098 triplets, the distractor contains an incorrect noun.
More detailed statistics regarding the number of triplets concerning specific concepts are provided in Appendix \ref{sec:app:data_stats}.

\paragraph{A note on perceived gender annotations}
Our evaluation dataset uses annotations from the Open Images dataset, which rely on the physical appearance of persons to annotate their perceived gender. We use the provided annotations, and the resulting biases are unfortunately reproduced in our evaluation set. We discuss this issue in further detail in the Ethics Statement (Section \ref{sec:ethics}).

In \citet{salminen_inter-rater_2018} gender classification from face pictures by human annotators shows an inter annotator agreement greater than 95\%. True gender cannot be classified, and high inter-annotator agreement does not imply a correct gender choice, but we expect the gender annotations of Open Images to be reliable enough to be used as a basis for our analyses.

\subsection{Metric}

We evaluate pre-trained models on their image-text matching performance in a zero-shot setting, i.e. without any further training. For each triplet, we test whether the models give a higher similarity score for the correct sentence than for the distractor sentence. We calculate accuracy for each pair, i.e. the model needs to succeed for both the example and the counter-balanced example triplet.

For each pair of triplets $(t_1,t_2) = ([I_1, S_1, S_2],[I_2,S_2,S_1])$, we calculate the following score:
\[
f(t_1, t_2)= \begin{cases}
1, & \parbox{4cm}{if $s(I_1, S_1) > s(I_1, S_2)$ and $s(I_2, S_2) > s(I_2, S_1)$} \\
0, & \text{otherwise}
\end{cases}
\]
where $s(I, S)$ denotes the similarity between an image $I$ and a sentence $S$.
To obtain the similarity score, we use the softmaxed output of the image-text matching pretraining heads of the models.\footnote{For the model CLIP, we feed the image and both sentences at the same time, and obtain a similarity score for both sentences, where $s(I_1, S_1) = 1 - s(I_1, S_2)$.}

The final accuracy is the average score over all pairs in the evaluation set. Chance performance is at 25\%.\footnote{The model succeeds if the similarity scores fall into one of four possible configurations: $s(I_1, S_1) > s(I_1, S_2) \land s(I_2, S_1) > s(I_2, S_2) $; $s(I_1, S_1) < s(I_1, S_2) \land s(I_2, S_1) < s(I_2, S_2); s(I_1, S_1) > s(I_1, S_2) \land s(I_2, S_1) < s(I_2, S_2); s(I_1, S_1) < s(I_1, S_2) \land s(I_2, S_1) > s(I_2, S_2)$.}

As the dataset was manually filtered and requires only rather simple understanding of the images, we assume human performance to be close to 100\%. To verify this claim, we had a one person annotate a randomly sampled subset of 500 triplets. For each triplet, the annotator was asked to judge which of the two sentences describes the image better. The resulting performance was at 100\%.

\paragraph{A topline: the cropped task}
In order to explore the effect of the visual distractors on this noun-predicate dependency task, we additionally evaluate all models in a \texttt{cropped} task:  We reduce the image to the bounding box of the target object. Thus, the cropped image usually\footnote{If the bounding boxes of the target and visual distractor object overlap to a high degree, the cropped image might still contain (parts of) the distractor object.} only contains the target object, and no more visual distractors (i.e., the referent of the noun or the predicate in the distractor sentence is no longer present in the cropped image). To succeed at this (simpler) task, the model no longer needs to capture the predicate-noun dependency, it just needs to ground the single words correctly.
We use this task to estimate how much the performance of the models is affected by the ability to ground nouns and predicates in our evaluation dataset, in comparison to the (more sophisticated) ability of understanding predicate-noun \textit{dependencies}.

\subsection{Models}

We consider a range of state-of-the-art V\&L models that are pre-trained using text, image, and multimodal pretraining objectives on corpora of parallel image and text data. 
All models use the transformer architecture \cite{vaswani_attention_2017}, but vary in terms of pretraining data and objectives, image encoders, and multimodal fusion approaches.

In addition to their image and text pretraining objectives, the models commonly make use of an image-text matching objective, where the models are asked to predict whether a given sentence describes an image or not. We leverage the output of the corresponding pretraining head for calculating image-text similarities for our task.\footnote{The multimodal pretraining objective of Oscar does not involve matching and mismatching images and descriptions, but only matching and mismatching sequences of object tags. Therefore, we evaluate the checkpoint that has been fine-tuned for image-text retrieval.}

We evaluate LXMERT \cite{tan_lxmert_2019}, UNITER \cite{chen_uniter_2019}, ViLBERT \cite{lu_vilbert_2019}, Oscar \cite{li_oscar_2020}, VinVL \cite{zhang_vinvl_2021}, ViLT \cite{kim_vilt_2021}, and CLIP \cite{radford_learning_2021}. We could not evaluate the original VL-BERT \cite{su_vl-bert_2020}, because it was not pre-trained using an image-text matching loss. We also did not evaluate the original VisualBERT \cite{li_visualbert_2019}, as their implementation of the image-text matching loss requires one correct caption (in addition to a possibly faulty caption), which is not available for the Open Images dataset. Both models were however evaluated in the controlled conditions using VOLTA (see Section \ref{sec:results_volta}).

\paragraph{Pretraining datasets}

ViLBERT and VL-BERT are pretrained on Conceptual Captions \cite{sharma_conceptual_2018}. UNITER, LXMERT, ViLT, and VinVL\footnote{VinVL is pre-trained on parts of the Open Images dataset, and we found that most images used in our evaluation set are indeed part of the VinVL pretraining dataset. Because of this confound, results of VinVL are not directly comparable to the other models (even though we test the model here with novel sentences with respect to the images). Our results show however that VinVL compares very closely to Oscar, which has the same architecture, suggesting that the access to the images during training does not substantially affect the model's performance. The same confound is possibly present for CLIP, for which the training data is not public} make use of additional publicly available datasets such as COCO \cite{lin_microsoft_2014}, SBU captions \cite{ordonez_im2text_2011}, Flickr30K \cite{young_image_2014}, VisualGenome \cite{krishna_visual_2017}, and VQA datasets \cite{goyal_making_2017,zhu_visual7w_2016,hudson_gqa_2019}. The original VisualBERT is only pre-trained on COCO. Appendix \ref{sec:app:pretraining_datasets} details the pretraining datasets for all models as well as their sizes. The pretraining data for CLIP has not been publicly released, the authors state that it consists of 400M image-text pairs from the internet (an order of magnitude more data than for most of the other models, which do not surpass 10M image-text pairs in size).

\section{Results}

\subsection{Original Implementations}

We test all models using the evaluation methods and data described above. We make use of pre-trained models made publicly available by the authors.

Resulting accuracies are shown in Table \ref{tab:results}. We find that only some models perform substantially above chance, notably ViLT, UNITER and LXMERT. In the \texttt{cropped} task, performance is much higher for all models, with VinVL and ViLT reaching the highest performance.  This gap in performance between the \texttt{full} and \texttt{cropped} tasks indicates that while those models can match nouns and predicates in the image with the corresponding words rather well, they struggle to take into account the dependencies between them.

\begin{table}[htb]
    \centering
    \begin{tabular}{lcc}
\toprule
 & \multicolumn{2}{c}{Accuracy} \\
                  \cline{2-3}
  Model &                Full &      Cropped \\
\midrule
 LXMERT & $0.57$ &        $0.69$ \\
 UNITER & $0.54$ &        $0.64$ \\
ViLBERT & $0.28$ &        $0.66$ \\
   ViLT & $0.40$ &        $0.75$ \\
  Oscar & $0.32$ &        $0.67$ \\
  VinVL & $0.30$ &        $0.76$ \\
   CLIP & $0.20$ &        $0.59$ \\
\midrule
 Chance       & $0.25$ & $0.25$ \\

\bottomrule
\end{tabular}
    \caption{Accuracy of models trained in original conditions when provided the full images and when only exposed to the target object in the \texttt{cropped} task.}
    \label{tab:results}
\end{table}

\subsection{Controlled Training Conditions}\label{sec:results_volta}

We additionally evaluate models that are trained in controlled (and therefore more directly comparable) conditions as proposed in the VOLTA framework \cite{bugliarello_multimodal_2021}. In this setup, all models are trained on Conceptual Captions using the same pretraining objectives (masked language modeling, masked object classification, and image-text matching) and use the same image features, extracted from a Faster R-CNN.

We evaluate all models for which pretrained weights are available.
Resulting accuracy scores are presented in Table \ref{tab:results_volta}.

\begin{table}[htb]
    \begin{tabular}{lcc}
\toprule
                  & \multicolumn{2}{c}{Accuracy} \\
                  \cline{2-3}
            Model &                Full &      Cropped \\
\midrule
    CTRL\_UNITER & $0.24$ &        $0.63$ \\
    CTRL\_LXMERT & $0.20$ &        $0.56$ \\
   CTRL\_ViLBERT & $0.27$ &        $0.66$ \\
   CTRL\_VL-BERT & $0.24$ &        $0.66$ \\
CTRL\_VisualBERT & $0.20$ &        $0.64$ \\
\midrule
 Chance       & $0.25$ & $0.25$ \\
\bottomrule
\end{tabular}
    \caption{Accuracy of models trained in controlled conditions when provided the full images and when only exposed to the target object in the \texttt{cropped} task.}
    \label{tab:results_volta}
\end{table}

We find that under these controlled conditions, all models perform comparably and generally around chance level. It is therefore not straightforward to draw any conclusions regarding the effect of model architecture from these results.

In the \texttt{cropped} task, performance is much higher, with ViLBERT and VL-BERT reaching the highest performance. The performance gap between the two tasks (i.e., \texttt{full} vs. \texttt{cropped}) is substantially larger than for the original implementations, suggesting that the models are even less sensitive to predicate-noun dependencies under these controlled training conditions.

%ViLBERT performs similarly compared to the controlled conditions. This could be explained by the fact that this implementation is only trained on Conceptual Captions, like in the controlled setup.

\section{Analyses and Discussion}

\subsection{Comparing Model Performances}

\paragraph{The role of pretraining data}

Within the set of the evaluated models, we do not find evidence for a correlation between the size of the pretraining dataset and the model's ability to capture predicate-noun dependencies (see also Appendix \ref{sec:app:pretraining_datasets}).
Despite being trained on comparable or even larger amounts of data, ViLT, Oscar and VinVL perform substantially worse than LXMERT and UNITER. CLIP performs below chance level, despite having by far the largest pretraining dataset.

The pretraining data of CLIP is not publicly available, but as it was automatically scraped from the internet we believe the quality (i.e descriptiveness) of its captions to be comparable to that of Conceptual Captions. In additional experiments (see Appendix \ref{sec:app:clip_additional_results}), we study the performance of CLIP models trained on different datasets using a range of publicly available model checkpoints. The performance of CLIP remains below chance level for all tested checkpoints. This might be because all available checkpoints are all trained on rather noisy data, or because the architecture and pretraining objectives of CLIP don’t allow it to learn grounded predicate-noun dependencies.

Datasets that are composed of highly descriptive captions seem to be advantageous for the learning of noun-predicate dependencies.
Indeed, for datasets such as COCO \cite{lin_microsoft_2014} or VQA \cite{antol_vqa_2015}, the images are not only strongly associated with the captions or question–answer pairs (as they were crowdsourced specifically for the tasks), but also precise and detailed in nature.
In contrast, Conceptual Captions \cite{sharma_conceptual_2018} is composed of images with captions that were automatically collected from web pages, and therefore generally rather broad descriptions of the image content.

ViLBERT and models trained in the controlled conditions are only trained using Conceptual Captions, and the resulting performances are around chance level. UNITER and LXMERT perform much worse compared to their original training setups. One main difference for these two models in their original implementation compared to the controlled condition is that they are trained on richer datasets with respect to the language modality, leveraging more descriptive captions.\footnote{The original pretraining datasets for UNITER and LXMERT are also larger in terms of the number of image-text pairs. However, LXMERT is actually trained on much fewer unique images than in the controlled conditions (180K vs. 3.1M). Therefore, we assume that the sheer size is not the driving factor of performance.} %\footnote{Another difference that could cause the performance drop is the choice of pretraining objectives, as discussed in the upcoming paragraph.}

This observation is coherent with what \citet{hendricks_probing_2021} found when studying verb understanding of V\&L models: They compare performance of the same model when trained on Conceptual Captions or COCO, and find that the model trained on COCO performs better, despite Conceptual Captions being bigger and closer to the task in terms of image and language distribution. 

These results suggest that, when considering multimodal dependencies, having a high quality pretraining dataset with less noise and more descriptive textual data could be more important than having a larger dataset. Highly descriptive textual data is essential to learn precise predicate-noun dependencies.

\paragraph{The role of pretraining objectives}

While models such as ViLT, Oscar, and VinVL are trained on datasets that are comparable in size and quality to those of LXMERT and UNITER, they still perform substantially worse on the task.
One explanation could be that contrary to the other models, UNITER and LXMERT both have multimodal pretraining objectives \textit{in addition} to image-text matching: Visual question answering for LXMERT and word-region alignment for UNITER.\footnote{ViLT also uses a word-patch alignment objective similar to word-region alignment, but the patches are not based on regions detected by an object detector and therefore the loss can not leverage any semantic labels for the patches during training, making this multimodal objective probably less useful.}
This could help the models to establish finer multimodal dependencies.
Indeed, ViLT and VinVL show better results than UNITER and LXMERT in the \texttt{cropped} task (indicating that their object/predicate recognition performance even surpasses that of the other models), but worse results in the \texttt{full} task. Our hypothesis is that the pretraining objectives of UNITER and LXMERT enable them to learn more fine-grained multimodal dependencies than ViLT and VinVL, even though their performance on the \texttt{cropped} task is worse.\footnote{Most directly comparable are probably the cases of ViLT and UNITER, which are both trained on the same datasets, but with different pretraining objectives.}

This gap in performance should not only be due to the training data associated with the additional pretraining objectives, as VinVL also uses data from Visual Question Answering task, but without training on the objective.

 %ViLT shows better results than UNITER in the \texttt{cropped} task and worse results in the \texttt{full} task. %Our hypothesis is that since ViLT does not have multimodal objectives other than image-text matching, it learned less fine-grained multimodal dependencies than UNITER, which used word-region alignment.

The impact of the multimodal pretraining objectives of UNITER and LXMERT can be an additional explanation for the drop in performance of CTRL\_UNITER and CTRL\_LXMERT, which were only trained using image-text matching as a multimodal pretraining objective. The gap in performance between those controlled models and the original models indicate that using more precise multimodal pretraining objectives and better annotated datasets can greatly improve the learning of multimodal dependencies.

The lack of suitable multimodal pretraining objectives could also offer an explanation for the poor performance of CLIP in our task.

\paragraph{The role of image encoders}

The authors of ViLT and VinVL motivate their work by suggesting that improved image features are mandatory for improved multimodal reasoning of V\&L transformers. Here, we observe that these improved features only translate to better results in the \texttt{cropped} task (where ViLT and VinVL perform best). We speculate that the improved image encoders allow for a better understanding of visual entities, but not necessarily of the dependencies between them. In order to obtain more conclusive interpretations regarding the role of image features, we require more targeted experiments which control for other confounding factors present here (such as different pretraining objectives).
%However, we could speculate that using both finer multimodal objectives and improved image features could lead to better overall results.

\paragraph{The role of model architecture}

In addition to the lack of suitable pretraining objectives, the worse performance of CLIP compared to the other models could also be due to the fact that it does not support any kind of inter-modal fusion of features within the model (image and text are processed in separate submodules that do not allow for inter-modal interaction).
This shortcoming of CLIP is also discussed in \citet{kim_vilt_2021}, where the authors find representations from CLIP to be not useful for the more advanced multimodal reasoning task NLVR2 \cite{suhr_corpus_2019}.

However, there seems to be no major effect of architecture with respect to multimodal fusion in the case of single and dual stream transformers: LXMERT and UNITER have comparable performances, even though one is dual-stream transformer and the other a single-stream transformer.

\subsection{Performance for Nouns vs. Predicates}

Here, we compare performance for pairs in which the sentences differ with respect to the noun, to sentences with a different predicate. Detailed results for all models are reported in Appendix \ref{sec:app:results_noun_predicate}.
Overall patterns show a slightly better performance for cases in which the noun was switched, especially in the \texttt{cropped} task. This is in line with findings that V\&L models are better at grounding nouns than verbs \cite{hendricks_probing_2021}.

\subsection{Analysis of individual nouns and predicates}\label{sec:per_concept_acc}

For a given concept (noun or predicate), we consider all pairs that contain this concept in at least one of the two sentences, i.e. cases in which a model's understanding of a concept is instrumental for making the correct decision.

\begin{figure}[htb]
    \centering
    \includegraphics[width=.5\textwidth]{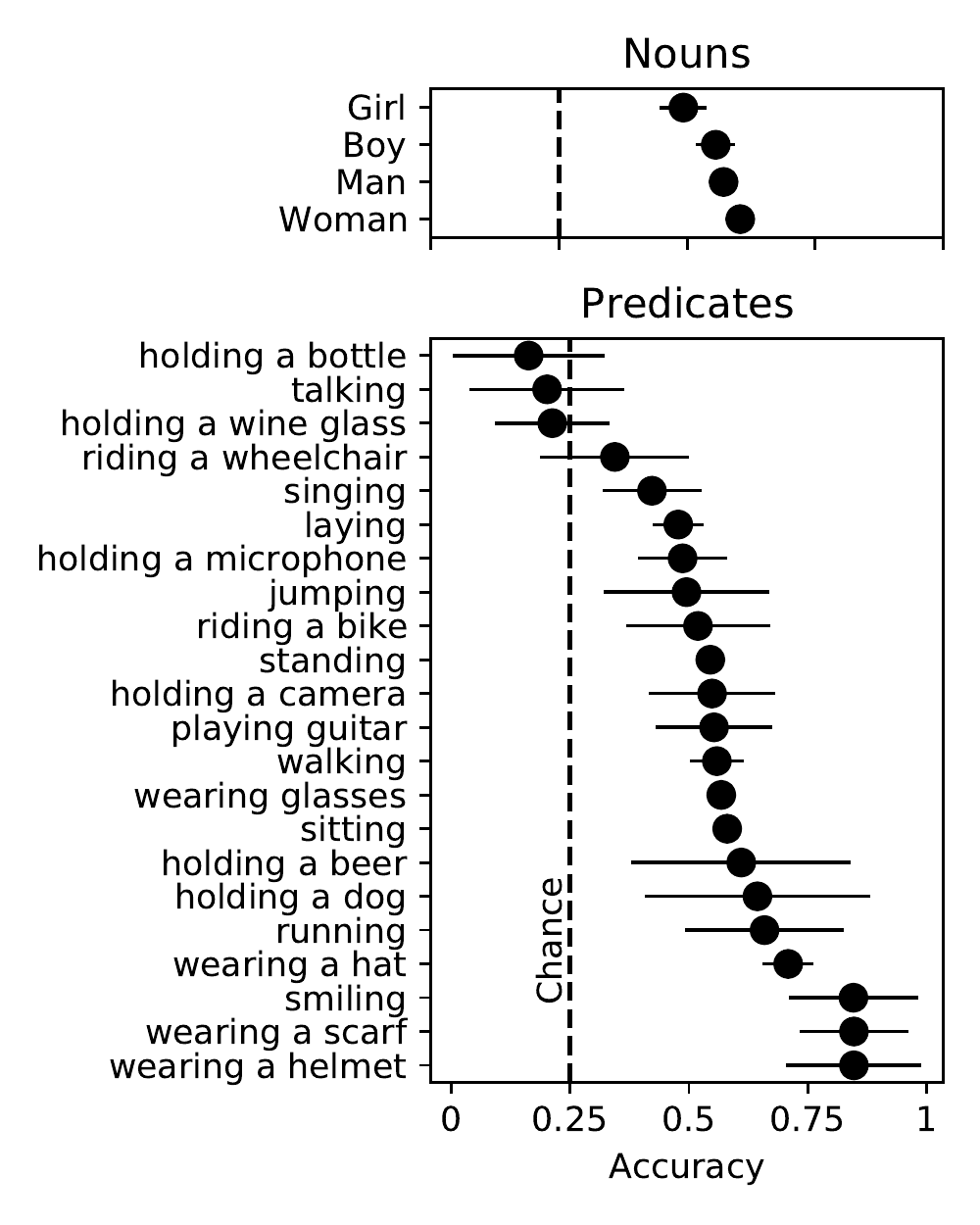}
    \caption{Per-concept accuracies for LXMERT. We display nouns and predicates for which we have at least 10 evaluation triplets. Standard deviation calculated using bootstrapping (100 re-samples).}
    \label{fig:accuracies_per_concept_lxmert}
\end{figure}

Figure \ref{fig:accuracies_per_concept_lxmert} shows the per-concept accuracies of the best performing model, LXMERT. Appendix \ref{sec:app:acc_per_concept_all_models} shows the per-concept accuracies for all models in their original implementations.

We observe large variation in accuracy scores of predicates, and less variation for nouns. We could not find any simple reasons that explain the predicates' variability. For example, verbs can have good or bad performances (e.g., ``running'' vs ``talking''), and the same can be said for predicates that are composed of both verb and noun (e.g., ``holding a bottle'' vs. ``wearing a helmet''). That said, factors that may influence model performance on specific nouns or predicates are further discussed in Section \ref{sec:confounding_factors}.

Additionally, we observe that for some concepts, the models perform better if the concept is the target, and for others, performance is better if it is the distractor. This is, e.g., the case for the pair ``sing'' vs ``stand'', where the models consistently perform better if ``sing'' is the target predicate. Appendix \ref{sec:app:results_per_tuple} shows the accuracy for each (target, distractor) concept tuple.

\subsection{Confounding Factors}\label{sec:confounding_factors}

Here we discuss possible factors influencing the models' performances.

\paragraph{Object salience}

In most of the images in our evaluation set, the target and distractor persons in the image are not of equal size, and not equally salient (sometimes one is more in the foreground than the other). We explore whether there is an effect on the models' decisions by correlating the models' predictions with target and distractor bounding box size and location.

More specifically, we measure the difference in similarity for target and distractor sentence $s(I_1,S_1) - s(I_1,S_2)$ and correlate it with the difference in bounding box size of the target and distractor object. Further, we also correlate it with the difference of distances from the center of the image.

For LXMERT, we find no significant correlation (Bounding box size: Pearson $r= -0.03, p = 0.16$, bounding box distance from center: Pearson $r= 0.03, p = 0.14$). Correlation scores for other models can be found in Appendix \ref{sec:app:correlations}. While there are statistically significant correlations for some models, these are small and of varying direction. The largest correlations are found for CLIP (Bounding box size: Pearson $r= 0.14, p < 0.01$, bounding box distance from center: Pearson $r= -0.24, p < 0.01$), indicating that the performance of CLIP could be affected, to some extent, by object salience.

\paragraph{Concept recognizability}

We also correlate the models' similarity judgments differences to differences in concept recognizability, which we operationalize by taking the object or attribute confidence score for a given concept in an image from a Faster R-CNN \cite{ren_faster_2015} trained on VisualGenome.\footnote{If there are multiple objects/attributes with the corresponding label in the image, we take the maximum confidence.}

For most models, we find a small positive correlation (see Appendix \ref{sec:app:correlations}), indicating that the models' similarity judgments are affected by the varying degree to which the concepts are recognizable in the image.

\paragraph{Linguistic biases}
Another aspect, already mentioned earlier, is that models' performance could be affected by linguistic biases in the training data, such as the frequency and co-occurrence of words and phrases.

To explore this possible effect, we correlate the difference in similarity for target and distractor sentence with the difference of target and distractor sentence perplexity. We calculate the perplexity for each sentence using a single-modality BERT model (\texttt{bert-base-uncased}), that was fine-tuned for 3 epochs on the textual data of Conceptual Captions. 

For LXMERT, we find no significant correlation (Pearson $r= -0.01, p = 0.48$). For the other models, we find very small positive correlations (see Appendix \ref{sec:app:correlations}).
We conclude that the models do not rely only on shallow heuristics of the training data in the textual modality.

\section{Conclusion}
This work examines whether state-of-the-art V\&L models learn multimodal syntactic dependencies, by focusing on a case study on simple predicate-noun dependencies. Our controlled experiments and analyses on a range of recent models reveal that their capability track such dependencies is variable, with some models (e.g., LXMERT and UNITER) show performance above chance level and others (e.g., CLIP) performing even below chance.

In contrast to the recent trend in the field focused on increasing pretraining data and using simple general-purpose pretraining objectives \cite{brown_language_2020,devlin_bert_2019}, here we observe that best performance is achieved, rather, with high-quality pretraining data, and more fine-grained pretraining objectives.

More specifically, our results suggest that multimodal pretraining objectives have a major impact on the model's learning of grounded predicate-noun dependencies. Models that include more targeted objectives such as visual question answering and word region alignment in addition to the general image-text matching objective show better performance. 
%Models that leverage more recent image encoders do not perform better on the task.
In addition, having highly descriptive pretraining datasets seems to help with learning fine-grained multimodal dependencies. In comparison, models trained on larger, web-scraped datasets do not perform well.
%We did not find evidence for a significant effect of the models' architectures on performance. 

In the future, the proposed highly-controlled evaluation protocol can be used to conduct more targeted studies regarding the role of model architecture, pretraining objectives, as well as training data quality and quantity in order to build V\&L models that are better at learning grounded predicate-noun dependencies, and possibly also other, move advanced multimodal reasoning tasks. 

%One limitation in our study is that the \textsc{cropped} task is not really an optimal control task. It could be cropping out essential parts, and the image size is reduced.

%Compared to other adversarial evaluation tasks such as winoground \cite{thrush_winoground_2022} our task is arguably much easier (for a human), still the state-of-the-art models' performances are far from perfect.

\section{Limitations}
While our analyses revealed patterns that seems to explain observed variability in the models' performance, the role of some architectural choices such as image encoding techniques remains ambiguous.
A better understanding of all factors influencing the learning of grounded predicate-noun dependencies could be achieved by training sets of models on comparable conditions and by varying only one factor at a time \cite[as done for example in][regarding the role of model architecture]{bugliarello_multimodal_2021}.

The range of concepts evaluated is rather small and therefore not representative for the understanding of grounded predicate-noun dependencies in general. More targeted data collection will be necessary in order to obtain more large-scale evaluation datasets.
Additionally, our zero-shot evaluation paradigm introduces a possible mismatch between training and evaluation: Models are trained using pairs of images and descriptions where the descriptions often describe \textit{all} salient parts of the image, whereas in our evaluation set the descriptions focus on only \textit{one} aspect/person in the image.

In the \texttt{cropped} condition, the images are not are not representative of the typical photographic framing of image-text corpora, which could deteriorate our results. That said, random cropping is a frequent data augmentation technique in computer vision research, where it has been successfully applied to improve generalization performance \cite{krizhevsky_imagenet_2012}.

Further, some scenes, actions and cultures are disproportionally represented in our evaluation dataset. As proposed in \cite{liu_visually_2021}, it is important to pursue further work on more diverse datasets.

\clearpage
\section{Ethics Statement}\label{sec:ethics}

The proposed evaluation set relies on subjective annotations of perceived gender. 
Attempting to classify gender based on physical appearance is an ill-posed problem (e.g., due to limitations of object detectors, biases of the human annotators). Additionally, the annotations only consider \textit{binary} gender classes (woman/man, girl/boy). Algorithms that perform such classifications are neither ideal nor desirable, as they perpetuate harmful stereotypes and exclude non-binary gender identities  \cite[e.g.,][]{dev_harms_2021,hamidi_gender_2018,blodgett_language_2020,bender_dangers_2021}. We explored whether it would be possible to use other classes, but we did not find many examples that would allow for an evaluation of sensitivity to predicate-noun dependencies in a controlled fashion. As our image selection is very constrained (we require a visual distractor, and a counter-example image with reverse properties), we found only sufficient examples for the categories used in the paper. We initially started a bottom-up data exploration in which we considered all labels present in the Open Images dataset, but found only very few examples for a few other categories (generally less than 5 examples after manual filtering). This might be due to the focus of the annotations in Open Images, future work could explore the use of other datasets that are focused on other types of annotations, the main challenge being the requirement for sufficiently large datasets in order to find matching examples and counter-examples.
Future efforts should be dedicated to creating datasets that aim at more inclusive annotations.

We acknowledge the severity of these issues, and emphasize that our work does not promote applications of gender classification in downstream tasks, but only uses it as a basis for analysis of existing models.

\section{Acknowledgements}
We thank the anonymous reviewers for their useful comments and feedback.

Research supported by grants ANR-16-CONV-0002 (ILCB), ANR-11-LABX-0036 (BLRI), ANR-21-CE28-0005-01 (MACOMIC), AMX-19-IET-009 (Archimedes Institute) and the Excellence Initiative of Aix-Marseille University (A*MIDEX).
This work was performed using HPC resources from GENCI–IDRIS (Grant 2021-[101693]).

\bibliography{references,custom}
\bibliographystyle{acl_natbib}\balance

\clearpage
\appendix

\nobalance
\section{Appendix}

\subsection{Image Pre-Filtering}\label{sec:app:pre_filtering}

We consider labels that occur at least 100 times in the dataset.

As some labels are similar and sometimes used interchangeably by the annotators, we create groups of synonyms for some labels and treat labels within a group as identical in the following. The groups of synonyms can be found in Table \ref{tab:synonyms}.

Further, we verify that the bounding boxes of the target and distractor objects are big enough (at least 20\% width and 20\% height of the image) and that the bounding box sizes of target and distractor objects don’t differ by more than a factor of 2.

Finally, we ensure that there is at least one counter-example for each triplet before starting the manual image selection phase.

\begin{table}[hb]
    \centering
    \begin{tabular}{p{.46\textwidth}}
    ``Table'', ``Desk'', ``Coffee table'' \\
    ``Mug'', ``Coffee cup'' \\
    ``Glasses'', ``Sunglasses'', ``Goggles'' \\
    ``Sun hat'', ``Fedora'', ``Cowboy hat'', ``Sombrero'' \\
    ``Bicycle helmet'', ``Football helmet'' \\
    ``High heels'', ``Sandal'', ``Boot'' \\
    ``Racket'', ``Tennis racket'', ``Table tennis racket'' \\
    ``Crown'', ``Tiara'' \\
    ``Handbag'', ``Briefcase'' \\
    ``Cart'', ``Golf cart'' \\
    ``Tree'', ``Palm tree'' \\
    ``Football'', ``Volleyball (Ball)'', ``Rugby ball'', ``Cricket ball'', Tennis  ball'' \\
    \end{tabular}
    \caption{Groups of label synonyms. Each line corresponds to one group.}
    \label{tab:synonyms}
\end{table}

\subsection{Dataset statistics}\label{sec:app:data_stats}

\begin{figure}[htb]
    \centering
    \includegraphics[width=.5\textwidth]{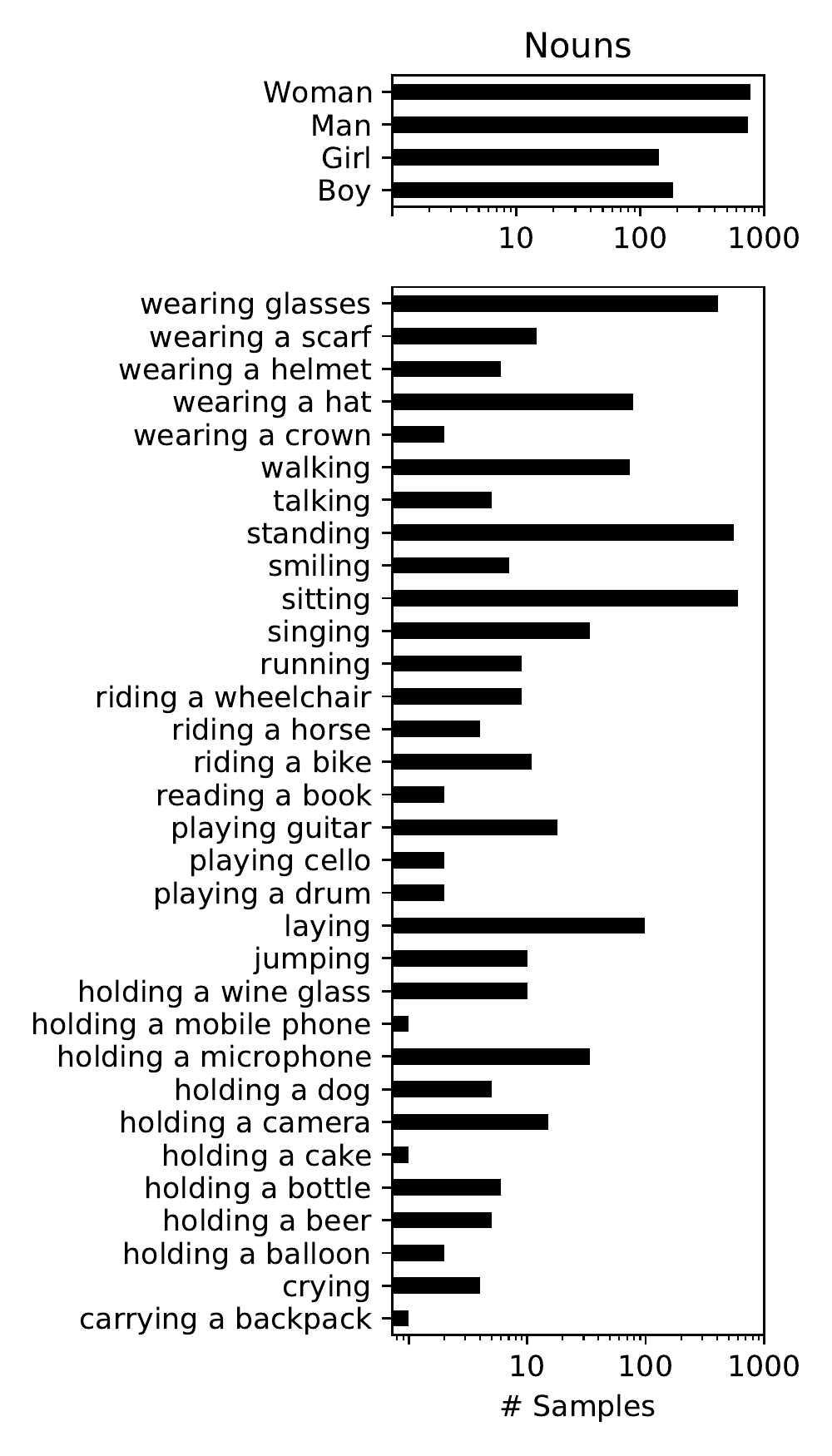}
    \caption{Number of triplets per concept. Note that the x-axis is on logarithmic scale.}
    \label{fig:num_samples_concepts}
\end{figure}

Figure \ref{fig:num_samples_concepts} shows the number of triplets for each noun and predicate. For a given noun or predicate, we count all pairs that contain this concept in at least one of the two sentences, i.e. cases in which correct understanding of a concept is useful for making the correct decisions. 

\subsection{Details on V\&L Models}\label{sec:app:pretraining_datasets}
\subsubsection{Pretraining Datasets}
\begin{table*}[ht]
    \centering
    \begin{tabular}{l|ccccccccc}
    & & & & & & & & \multicolumn{2}{c}{Total size} \\
    \cline{9-10}
    Model & CC & COCO & SBU & VG & QA & F30K & OI & \# images & \# image-text pairs \\
    \midrule
    LXMERT &  & \checkmark &  & \checkmark & \checkmark & & & 0.18M & 9.18M \\
    UNITER & \checkmark & \checkmark & \checkmark & \checkmark & & & &  4.16M & 9.59M \\
    ViLBERT & \checkmark &  &  &  &   & & &  3.10M & 3.10M \\
    ViLT & \checkmark & \checkmark & \checkmark & \checkmark & & & & 4.05M & 9.85M \\
    Oscar & \checkmark & \checkmark & \checkmark &  & \checkmark & \checkmark & & 4.10M & 6.50M \\
    VinVL & \checkmark & \checkmark & \checkmark &  & \checkmark & \checkmark & \checkmark & 5.65M & 8.85M\\
   % VisualBERT &  & \checkmark &  &  &  \\
   % VL-BERT & \checkmark &  &  &  &  \\
    \end{tabular}
    \caption{Pretraining datasets of the tested models: CC \cite[Conceptual Captions;][]{sharma_conceptual_2018}, COCO \cite{lin_microsoft_2014}, SBU captions \cite{ordonez_im2text_2011}, VG \cite[Visusal Genome;]{krishna_visual_2017}, F30K \cite[Flickr 30K;][]{young_image_2014}, OI (Open Images), and QA (including VQA2.0 \cite{goyal_making_2017}, GQA \cite{hudson_gqa_2019}, and VG-QA \cite{zhu_visual7w_2016}).}
    \label{tab:pretraindata}
\end{table*}
%https://docs.google.com/spreadsheets/d/1moz5O5FQsBMG9-lItf8Kv4QHqcs9jWGfiq39TibqZXQ/edit?usp=sharing

Table \ref{tab:pretraindata} details the multimodal datasets used for V\&L models.
Dataset sizes as reported in the corresponding papers. Note that these sizes are also affected by the fact that some models leverage validation sets for pretraining, while others constrain the data to the training sets. Also, different approaches for dataset overlap detection have been applied.
The pretraining data size for CLIP is reportedly 400M image-text pairs.

\subsubsection{Number of parameters}

\begin{table}[ht]
    \centering
    \begin{tabular}{l|r}
    Model & \# Parameters \\
    \midrule
    LXMERT & 228,051,752 \\
    UNITER & 112,938,887 \\
    ViLBERT & 250,044,029 \\
    ViLT & 111,596,546 \\
    Oscar & 111,062,018 \\
    VinVL & 111,686,973 \\
    CLIP & 151,277,313 \\
    \end{tabular}
    \caption{Number of parameters for each model.}
    \label{tab:num_of_parameters}
\end{table}
Table \ref{tab:num_of_parameters} compares the number of trainable parameters for each model that was tested in their original implementations.

\subsection{Additional Analyses}

\subsubsection{Results for CLIP with varying pretraining data}\label{sec:app:clip_additional_results}
Table \ref{tab:clip_additional_results} presents the accuracy scores of multiple publicly available checkpoints for CLIP trained on different training data. 

\begin{table}[htb]
    \centering
    \begin{tabular}{llc}
    \toprule
      Visual Encoder & Dataset    & Accuracy \\
    \midrule
    RN101 & YFCC-15M &    $0.18$ \\
    RN101 & 400M &        $0.21$ \\
    RN50 & cc12m &        $0.18$ \\
    RN50 & 400M &         $0.20$ \\
    RN50 & YFCC-15M &     $0.17$ \\
    ViT-B-32 & aion2b\_e16 &        $0.21$ \\
    ViT-B-32 & laion400m\_e31 &        $0.20$ \\
    ViT-B-32 & laion400m\_e32 &        $0.19$ \\
    ViT-B-32 & 400M &                $0.20$ \\
    ViT-L-14 & 400M (336px) &        $0.20$ \\
    \bottomrule
    \end{tabular}
    \caption{Accuracy (Full) of CLIP models with varying visual encoders and pretraining data.\label{tab:clip_additional_results}}    
\end{table}

\subsubsection{Controlling for linguistic robustness}\label{sec:app:linguistic_robustness}
As the sentences used in our evaluation dataset are built from a template, they do not vary in syntax. We verify that results obtained do not depend on the exact template chosen.

We vary the original templates by using the definite article (``the'') at the beginning of sentences, and using verbs in simple present instead of present progressive tense (e.g., ``the woman sits.'' or ``the man holds a camera.'').

Results with these alternative sentences are show in Table \ref{tab:results_alt_sentences}. We find that overall result patterns are highly similar to those with the original sentences in Table \ref{tab:results}.

\begin{table}[htb]
    \begin{tabular}{lcc}
\toprule
 & \multicolumn{2}{c}{Accuracy} \\
                  \cline{2-3}
  Model &                Full &      Cropped \\
\midrule
 LXMERT & $0.55$ &        $0.70$ \\
 UNITER & $0.54$ &        $0.66$ \\
ViLBERT & $0.26$ &        $0.67$ \\
   ViLT & $0.34$ &        $0.72$ \\
  Oscar & $0.32$ &        $0.65$ \\
  VinVL & $0.30$ &        $0.74$ \\
   CLIP & $0.21$ &        $0.58$ \\
\bottomrule
\end{tabular}

    \caption{Results with alternative sentences.}
    \label{tab:results_alt_sentences}
\end{table}

\subsubsection{Switching noun vs. switching predicate}\label{sec:app:results_noun_predicate}

Table \ref{tab:results_noun_predicate} presents the accuracy for cases in which target and distractor sentence differ with respect to the predicate, or noun. We report scores for all models in their original implementations.

\begin{table*}[htb]
    \centering\begin{tabular}{@{\extracolsep{4pt}}lcccc@{}}
\toprule
 & \multicolumn{2}{c}{Full} & \multicolumn{2}{c}{Cropped} \\
                  \cline{2-3} \cline{4-5}
  Model &              Noun &         Predicate &    Noun & Predicate \\
\midrule
LXMERT & $0.60$ &    $0.55$ &         $0.78$ &              $0.62$ \\
 UNITER & $0.60$ &    $0.50$ &         $0.76$ &              $0.56$ \\
ViLBERT & $0.27$ &    $0.28$ &         $0.74$ &              $0.59$ \\
   ViLT & $0.44$ &    $0.37$ &         $0.80$ &              $0.72$ \\
  Oscar & $0.36$ &    $0.30$ &         $0.75$ &              $0.62$ \\
  VinVL & $0.33$ &    $0.28$ &         $0.83$ &              $0.71$ \\
   CLIP & $0.21$ &    $0.19$ &         $0.69$ &              $0.52$ \\
\bottomrule
\end{tabular}
    \caption{Accuracy of models for cases in which the distractor sentence contains a different noun, or a different predicate.}
    \label{tab:results_noun_predicate}
\end{table*}

\subsubsection{Analysis of individual nouns and predicates for all models}\label{sec:app:acc_per_concept_all_models}

\begin{figure*}[htb]
    \centering
    \includegraphics[width=\textwidth]{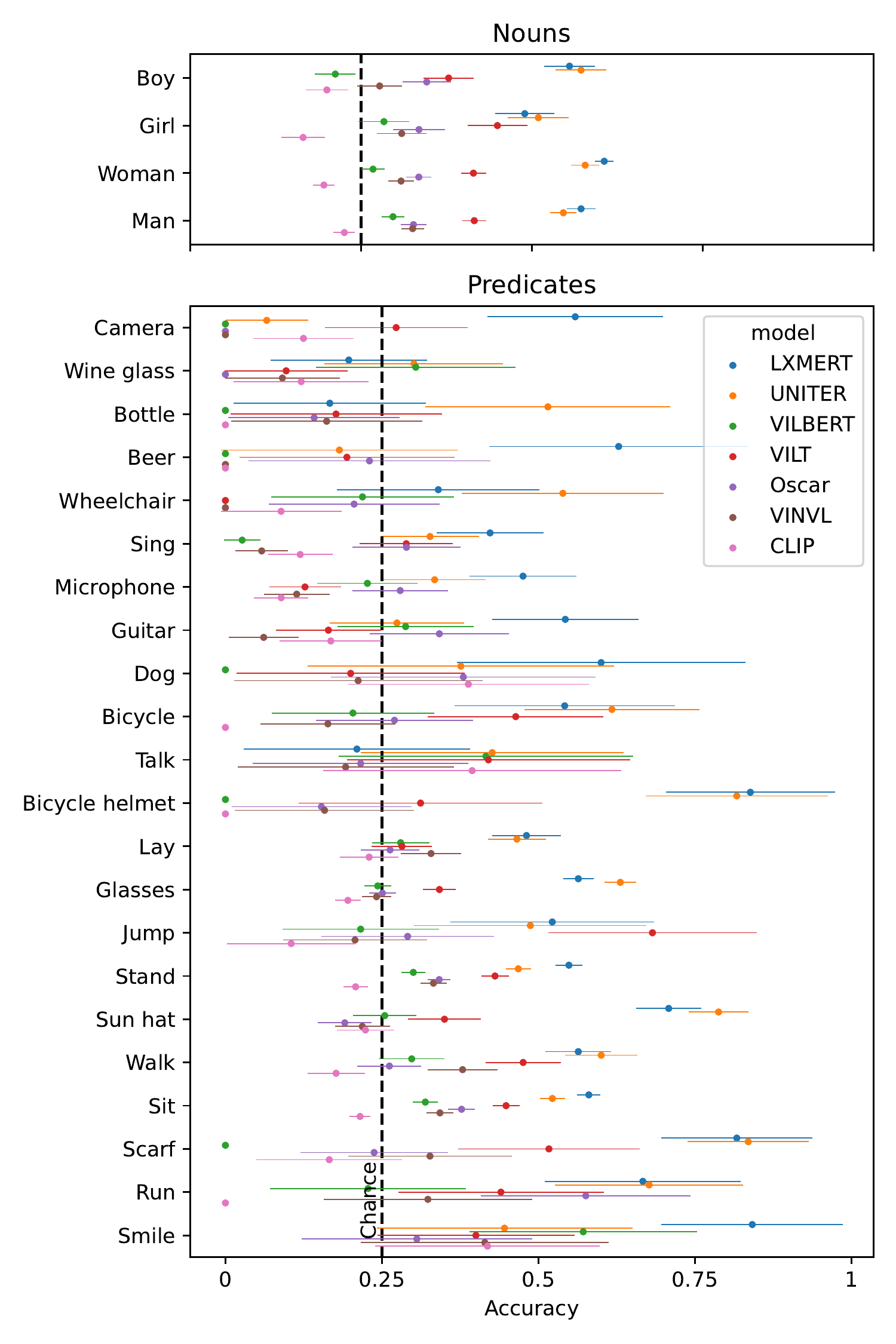}
    \caption{Per-concept accuracies for all models. We display nouns and predicates for which we have at least 10 evaluation triplets. Standard deviation calculated using bootstrapping (100 re-samples).}
    \label{fig:accuracies_per_concept}
\end{figure*}

Figure \ref{fig:accuracies_per_concept} shows the accuracies for split up for the different predicates and nouns for all models. For more details, refer to Section \ref{sec:per_concept_acc}.

\subsubsection{Accuracies for (target, distractor) tuples}\label{sec:app:results_per_tuple}

\begin{figure*}[htb]
    \centering
    \includegraphics[width=\textwidth]{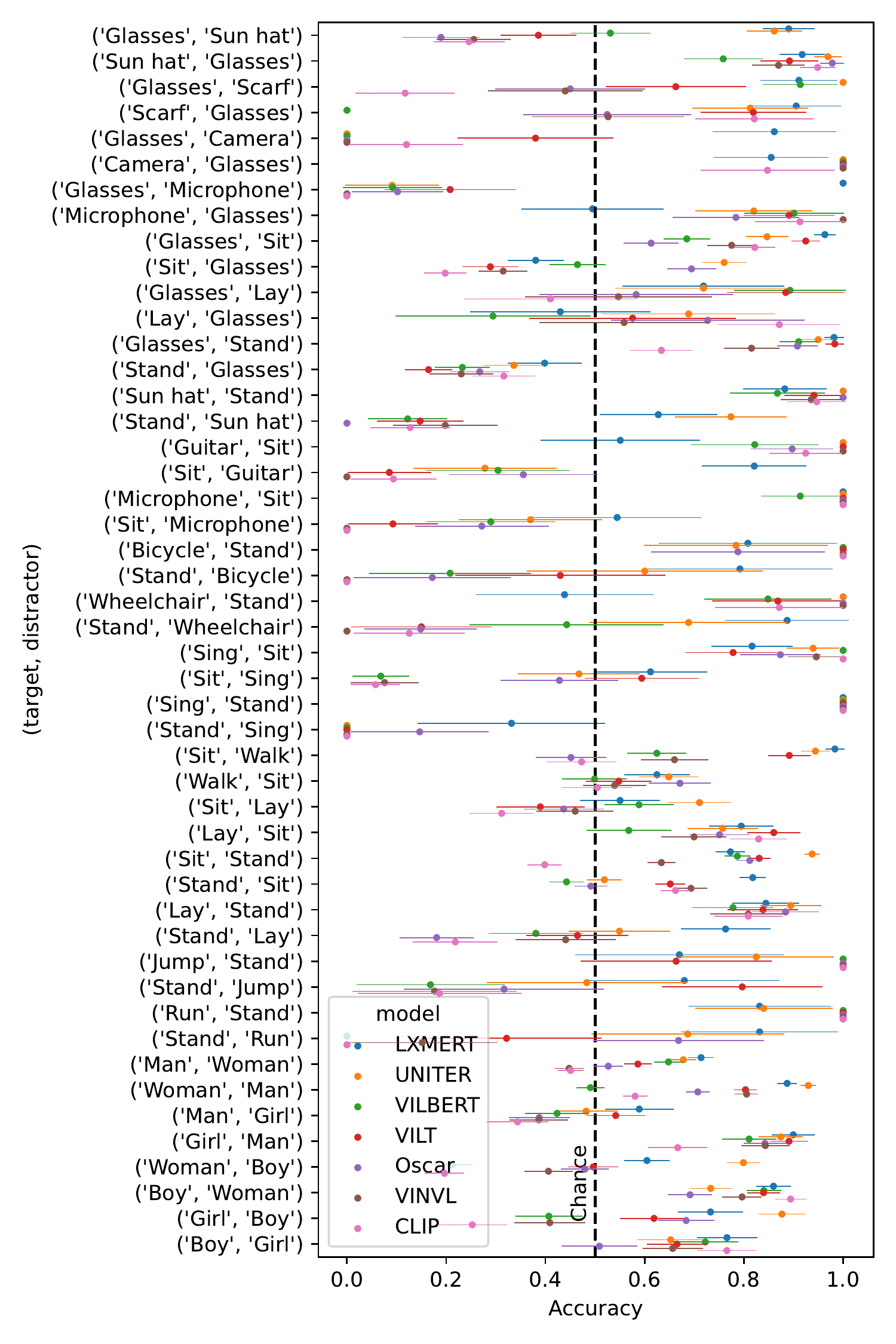}
    \caption{Accuracy for (target, distractor) tuples for all models. We display tuples for which we have at least 10 evaluation triplets. Note that chance performance is at $0.5$, because we report per-triplet (and not per-pair) accuracy.}
    \label{fig:accuracies_target_distractor}
\end{figure*}

Figure \ref{fig:accuracies_target_distractor} shows the accuracy for target-distractor tuples for all models.

\subsubsection{Confounding Factors}\label{sec:app:correlations}

In Table \ref{tab:correlations} we show the correlation scores for several confounding factors as described in Section \ref{sec:confounding_factors}.

\begin{table*}[htb]
    \centering\begin{tabular}{llllll}
\toprule
  Model &   Bounding box size & Distance from center &       Perplexity & Object detector confidence \\
\midrule
 LXMERT & -0.03 (p=0.12) &       0.03 (p=0.08) & -0.01 (p=0.48) &   0.30 (p=0.00) \\
 UNITER & -0.09 (p=0.00) &       0.09 (p=0.00) &  0.05 (p=0.01) &   0.26 (p=0.00) \\
ViLBERT &  0.11 (p=0.00) &      -0.16 (p=0.00) &  0.05 (p=0.02) &   0.22 (p=0.00) \\
   ViLT & -0.01 (p=0.73) &       0.03 (p=0.13) &  0.05 (p=0.02) &   0.26 (p=0.00) \\
  Oscar &  0.12 (p=0.00) &      -0.17 (p=0.00) &  0.06 (p=0.00) &   0.15 (p=0.00) \\
  VinVL &  0.12 (p=0.00) &      -0.12 (p=0.00) &  0.05 (p=0.01) &   0.04 (p=0.05) \\
   CLIP &  0.14 (p=0.00) &      -0.24 (p=0.00) &  0.08 (p=0.00) &   0.17 (p=0.00) \\
\bottomrule
\end{tabular}

    \caption{Correlations between difference in similarity and various factors related to targets and distractors: difference in bounding box size, distance from the image center of the bounding boxes, perplexity, and confidence scores of the bounding box as calculated using a Faster R-CNN object detector model.}
    \label{tab:correlations}
\end{table*}

\end{document}